\documentclass[10pt,twocolumn]{article}
\usepackage[T1]{fontenc}
\usepackage[utf8]{inputenc}
\usepackage{booktabs}
\usepackage{amsmath}
\usepackage{graphicx}
\usepackage{graphicx}
\usepackage{pgfplots}
\usepackage[font=small,labelfont=bf,labelsep=endash,margin=1.6em,skip=5pt]{caption}
\usepackage{float}   %
\pgfplotsset{compat=1.17}
\usetikzlibrary{positioning}
\usepackage[margin=20mm]{geometry}
\usepackage[round]{natbib}
\usepackage{url}
\usepackage{amssymb}   %
\usepackage{tipa}      %
\usepackage{gb4e}      %
\noautomath            %
\usepackage[hidelinks]{hyperref}  %
\newcommand{\gascore}[1]{\textsc{#1}}

\title{A POS Tier Is the Key to Automated Annotation for Low-Resource
Language Documentation: Neural Interlinear Glossing for Irabu,
a Southern Ryukyuan Language}
\author{{\Large Michinori Shimoji}\\[3pt]Kyushu University\\\texttt{smz@kyudai.jp}}
\date{}

\begin{document}
\maketitle

\begin{abstract}
Discourse data are the primary empirical basis of grammar writing in field
linguistics, but producing interlinearized text is notoriously expensive ---
on the order of one hour of work per minute of recording.  For endangered
languages, where the time remaining to verify analyses with native speakers is
itself limited, automating parts of the interlinearization workflow has
direct documentary value.  We implement a full neural annotation pipeline
(morpheme segmentation $\rightarrow$ POS tagging $\rightarrow$ glossing) for
Irabu Ryukyuan using deliberately small, transparent BiLSTM--CRF models, and
evaluate it under a realistic hard constraint: approximately one hour of fully
annotated discourse as the entire supervised resource.  Two factors of the
annotation itself are manipulated: its \emph{richness} (with or without a POS
tier) and its \emph{quantity} (training budgets from 6 to 47 minutes).  Gold POS improves grammatical
glossing by $+4.4\pm0.7$ points (significant in all 5 seeds), and the gain
grows as data shrink ($+11.6$ points at a quarter of the data); a POS tier
more than halves the amount of glossed data needed to reach a given accuracy.
In a fully automatic pipeline this gain is not yet realized: the tagger still
errs on 12\% of morphemes, and an incorrect POS misleads the glossing model
more than no POS at all.  The value is latent rather than lost: degrading gold
POS with controlled noise shows the gain returning as tagger accuracy rises,
with break-even near our tagger's current 88\% and $+1.6$ to $+3.2$ points
recovered at 92--96\%.
We conclude with a concrete recommendation for documentation practice:
annotate quadrilinearly --- text, POS, gloss, translation.
\end{abstract}

\medskip

\section{Introduction}
\label{sec:intro}

\paragraph{Background: the annotation bottleneck.}
Documentary linguistics takes as its primary object a corpus of naturalistic
communicative events --- narratives, conversations, and whatever further
event types the community's culture-specific ways of speaking comprise: a
``systematics'' of communicative events \citep[pp.~176--177]{himmelmann1998}.  For grammar writing, such connected discourse is an
indispensable empirical resource: it supplies the frequency, context, and
construction diversity that elicitation alone cannot
\citep{shimoji2011,shimoji2021}.
The cost side is equally well known: the gap between what is recorded and
what is transcribed --- the ``transcription bottleneck''
\citep[p.~e325]{seifart2018} --- reflects, in our experience, on the order of
\emph{one hour of expert work per minute of recording} for fully
interlinearized text (\citealp{shimoji2011}; a similar guess is given by
\citealp[p.~e336]{seifart2018}).  In documentation projects for endangered languages,
the Japonic--Ryukyuan languages among them, this annotation cost is the
standing bottleneck between \emph{recording} discourse and actually
\emph{using} it.
For endangered languages the bottleneck is compounded by a harder limit: the
time remaining with fluent speakers.  Every hour spent mechanically filling in
glosses is an hour not spent checking analyses \emph{with} speakers.
Research on automatic annotation is therefore not a convenience but a
necessity for documentary linguistics, and the present study is squarely in
that context: automation not as a replacement for the linguist, but as a way
to reallocate scarce speaker-facing time.

\paragraph{Stance: leverage, not replacement.}
The last point deserves emphasis, because it sets this study apart from much
of the current, largely NLP-driven discussion of low-resource annotation
automation, which tends --- often implicitly --- toward \emph{replacement}:
systems designed and benchmarked as if the goal were to produce interlinear
text without the linguist in the loop.  The tendency lives in the
operational format rather than in stated aims: the SIGMORPHON glossing
shared task, for instance, is prefaced by the remark that such methods
``can never fully replace the expertise of a dedicated documentary
linguist'' \citep[p.~186]{ginnetal2023} --- yet what the task measures is precisely
fully automatic production.
For endangered-language documentation we consider that framing not just
unrealistic but misdirected.
Consider what a chef wants from automation: not to arrive at the restaurant
and find every dish already cooked, but to be relieved of the shopping, the
cleaning, and the slow, monotonous prep work --- and to have even that done
under the chef's supervision, in a form the chef can inspect, because
whatever leaves the kitchen carries the chef's name.  The documentary
linguist is in the chef's position.  The discourse data being automated has a
specific consumer: it exists to be used by descriptive linguists for grammar
writing (as well as, of course, by language communities for revitalization
purposes, which we do not go into in this study), and its annotations are
meaningful only insofar as they conform to
the gold analysis that the linguist's own descriptive work produces --- the
label inventory, the segmentation conventions, the glossing decisions.
Automatic output that drifts from that analysis is not a cheaper version of
the same resource; it is a different resource of unclear value.  The
productive frame, we suggest, is the opposite of replacement: treat the small
body of training data that a descriptive linguist can carefully hand-annotate
in the time that remains as an \emph{asset}, and ask how to \emph{leverage}
that asset for the maximum documentary return.  The two questions below are
the operational form of this view.

\paragraph{Problem: two questions without concrete answers.}
Bringing automatic annotation into a documentation project immediately raises
two practical questions.  The first is \emph{model choice}: what kind of
model is appropriate under the data conditions of endangered-language
documentation?  The second --- asked less often, but the one a project leader
must answer first --- is \emph{training cost}: how much supervised data, of
what kind, must the linguist create before automation pays off?  That is,
the minimum \emph{quantity} and the minimum \emph{richness} of the teaching
annotation.  For Japonic--Ryukyuan languages, concrete studies addressing
either question are, to our knowledge, all but nonexistent
\citep[cf.][]{shimojiryukyu49}.  The gap has consequences well beyond
automation itself.  Discourse-collection projects for endangered Japanese
dialects are currently proliferating,\footnote{E.g.\ NINJAL's collaborative
project \emph{Endangered Languages and Dialects in Japan} (2016--2022) and
its successors, and the ongoing public release of the Agency for Cultural
Affairs' emergency dialect survey recordings
(\url{https://www.ninjal.ac.jp/events_jp/20220716a/}); see
\citet{hidaka2026} for a survey of current curation efforts.
} yet how much discourse to record is typically set either as an open-ended
aspiration (``as much as possible'') or by convention --- one hour, two
hours --- with no empirical basis for the number; and \emph{how} the
collected discourse should be annotated receives even less attention: in many
projects the recordings are transcribed and given a free translation only,
with no morpheme-level annotation at all \citep[e.g.\ the curation efforts
surveyed in][and the COJADS dialect corpus, \citealp{cojads}]{hidaka2026}.
Without answers to the two questions above, there is in fact no principled
way to set either the number of hours or the depth of annotation.

\paragraph{Purpose.}
This study takes up both questions, in different ways matched to their
nature.  The first --- model choice --- we answer \emph{by design}: small,
transparent, purely supervised BiLSTM--CRF models, on grounds set out in the
next paragraph.  The second --- training cost --- is the empirical subject of
the paper.  We implement and evaluate a complete neural interlinearization
pipeline --- morpheme segmentation, POS tagging, and glossing --- for Irabu,
a Southern Ryukyuan language \citep{shimoji2008,shimoji2017}, under a
scenario we consider realistic for most endangered-language projects:
\emph{the entire supervised resource --- the asset to be leveraged --- is
about one hour of fully annotated discourse} (774 utterance units, of which
the 620-utterance training portion corresponds to roughly 47 minutes;
\S\ref{sec:data}), with sub-budgets down to 6 minutes evaluated explicitly
(\S\ref{sec:datasize}).
The question is not ``how well can neural models gloss with abundant data''
but ``what return does a given investment of annotation actually buy --- and
how should that investment be allocated?''
Answering it yields what current practice lacks: an empirically grounded
criterion --- \emph{how many minutes of discourse to annotate, and with which
tiers} --- in place of effort targets and conventional hour counts.  Our
budget-centered evaluation culminates in a concrete proposal: a
\emph{two-stage} design in which a small first-stage corpus is annotated
richly, then leveraged to automate the rest (\S\ref{sec:conclusion}).

\paragraph{Model choice (the first question): small, supervised, and transparent.}
All components are BiLSTM--CRF sequence labelers
\citep{hochreiteretal1997,lafferty2001,lample2016}.  We deliberately do not use large
pretrained Transformer models \citep{vaswanietal2017,ginnetal2024}, for two
reasons.  First, the data reality: the premise of unsupervised or transfer
approaches --- large amounts of raw text --- does not hold for most endangered
languages, where the recorded discourse itself is minuscule; under such
conditions we consider it more realistic to invest the linguist's expertise
in \emph{careful annotation of the small data} and use it as direct
supervision.  We have previously framed this as a \emph{quality-dependent}
rather than \emph{quantity-dependent} scenario \citep{shimojiryukyu49}: an
experienced fieldworker can produce a small but highly consistent
gold-standard corpus, which is precisely the setting where small supervised
models are viable.  Second, workflow transparency: a BiLSTM--CRF factorizes
into per-morpheme evidence (emission scores) and label-sequence regularities
(transition scores), a structure that maps naturally onto how a linguist
reasons about interlinear annotation, and whose errors are correspondingly
diagnosable (\S\ref{sec:propagation} puts this transparency to work).  We
emphasize the scope of this choice: we make no claim that small supervised
models outperform pretrained ones --- indeed, pretrained multilingual encoders
can be viable even for low-resource languages
\citep{oguejietal2021,hangyaetal2022}, although ``low-resource'' there still
means raw text on the gigabyte scale or sizable monolingual corpora
(AfriBERTa pretrains on just under 1\,GB; the transcription tier of our
corpus totals $\sim$26\,KB, four orders of magnitude less) --- and
comparing against such encoders
and LLM-based approaches \citep{ginnetal2024,elsneretal2025,aycocketal2025}
under the same annotation budget is future work.  Our contribution concerns \emph{annotation design
under a fixed, simple model}.

\begin{figure*}[t]
\centering
\includegraphics[width=0.66\textwidth]{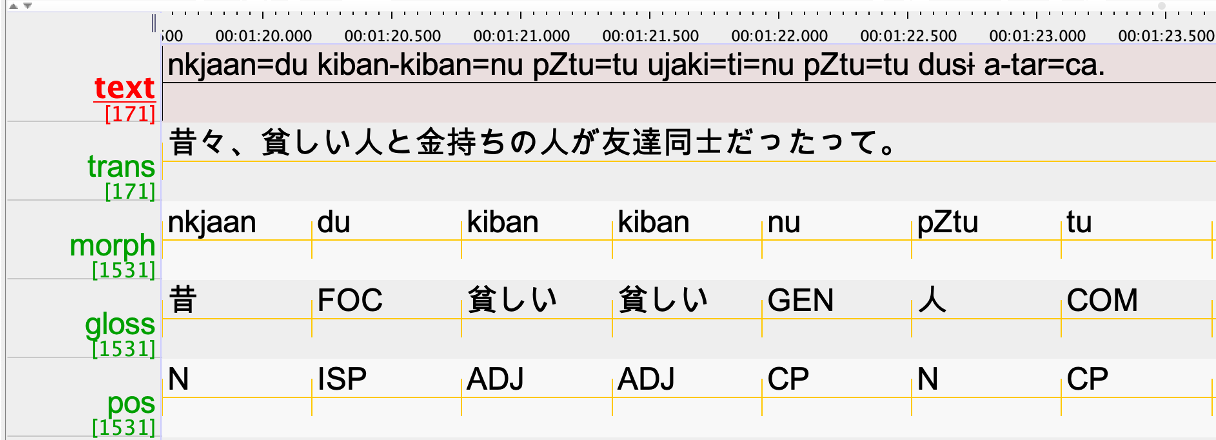}
\caption{Quadrilinear annotation in practice: the ELAN environment used to
build our corpus, showing the utterance \emph{nkjaan=du kiban-kiban=nu
pZtu=tu ujaki=ti=nu pZtu=tu dus\textbari{} a-tar=ca} `Once upon a time, (it
is said,) a poor man and a rich man were friends' (rendering of the
\texttt{trans} tier).  From top to bottom: \texttt{text} (word-segmented
transcription with \emph{-}/\emph{=} boundaries), \texttt{trans} (free
translation --- present in the corpus but \emph{not used} by our system),
\texttt{morph} (morpheme segmentation), \texttt{gloss} (morpheme-by-morpheme
glosses), and \texttt{pos} --- the tier that conventional trilinear practice
omits and that this paper argues for.}
\label{fig:elan}
\end{figure*}

\paragraph{Terminology: trilinear vs.\ quadrilinear.}
Standard IGT is \emph{trilinear}: a transcription line, a
morpheme-by-morpheme gloss line \citep{lehmann2004}, and a free-translation
line.  Note where the annotator's morpheme-level effort goes: it is
concentrated entirely in the \emph{gloss} line.  A parallel
\emph{morpheme-by-morpheme POS tier} is not part of common practice ---
part-of-speech information remains implicit in the analysis, or is recorded
only in the lexicon, but is rarely written down token by token.  We use
\emph{quadrilinear} for the format that adds exactly this tier (text + POS +
gloss + translation); Figure~\ref{fig:elan} shows the resulting tier
structure in our ELAN workflow.  Two clarifications are important.  First, the POS
tier at issue is \emph{per morpheme}, aligned with the gloss line, not a
per-word category label.  Second, while the free-translation line is part of
both formats, \emph{our system does not use it at any stage}: unlike
translation-leveraging approaches to glossing
\citep{zhaoetal2020,yangetal2024}, our pipeline operates entirely on the
transcription, POS, and gloss tiers.  Figure~\ref{fig:elan} shows what
quadrilinear annotation looks like in practice, in the ELAN environment in
which our corpus was built.  What our experiments quantify, therefore, is
the return on deliberately annotating the one morpheme-level tier that
standard practice omits.

\paragraph{What we test.}
Holding the model fixed, we manipulate the annotation itself along two axes:
\begin{itemize}
\item \textbf{Richness (RQ1):} does the POS tier --- quadrilinear rather
  than trilinear annotation --- improve automatic glossing?  We separate the
  intrinsic value of POS information (oracle input) from its realizable
  value in a full pipeline (predicted input), and dissect the gap token by
  token.
\item \textbf{Quantity (RQ2):} how does performance scale from 6 to 47
  minutes of annotated discourse, and does it interact with richness?
\end{itemize}

\paragraph{Contributions.}
(i)~A complete, reproducible neural interlinearization pipeline for a
low-resource Ryukyuan language,\footnote{All code (the full pipeline and
every experiment script), the gold corpus, and the complete per-sentence
test outputs reported in this paper are available at
\url{https://github.com/MichinoriShimoji/ML_autogloss}.
} trained on $\sim$47 minutes of annotated
discourse, with budgets down to 6 minutes evaluated (segmentation span-F1
0.907; POS accuracy 0.881; grammatical-gloss accuracy 0.93 with oracle POS).
(ii)~A controlled ablation showing that POS information improves grammatical
glossing by $+4.4\pm0.7$ points (all 5 seeds individually significant), with
the gain concentrated on linguistically predictable ambiguities (case/focus
clitics).
(iii)~An error-propagation analysis showing that with a realistic tagger the
gain is cancelled --- incorrect POS is \emph{worse than no POS} --- and an
exact decomposition of the loss, plus a noise-injection analysis locating the
break-even tagger accuracy ($\approx$88\%).
(iv)~A data-size analysis showing the POS gain \emph{grows} as data shrink
(up to $+11.6$ points at $\sim$12 minutes), so a POS tier more than halves
the glossed data needed --- exactly in the regime where documentation projects
begin.
(v)~A concrete recommendation for annotation practice: quadrilinear
annotation, with quantified costs and benefits --- and, building on it, a
\emph{two-stage design} for discourse-documentation projects (collect
$\sim$30 minutes and annotate it richly first; then collect more and let the
pipeline draft the annotations).
(vi)~A working implementation of the Stage-2 workflow: a browser-based
annotation tool in which the pipeline drafts the interlinearization of
arbitrary input, the annotator confirms or corrects every tier, and the
corrections feed back into the models (Figure~\ref{fig:tool}).

\section{Related Work}
\label{sec:related}

\paragraph{Automatic interlinear glossing.}
Automating IGT production has a substantial NLP literature.
\citet{palmer2009} evaluated machine-in-the-loop annotation strategies for
documentation; \citet{moelleretal2018} showed that CRF-based glossing reaches
useful accuracy for Lezgi from roughly 3{,}000 words of IGT --- a setting
close to ours in scale and spirit, down to the design decision of reducing
open-class stems to a placeholder and predicting the closed-class
grammatical glosses --- and \citet{barrigamartinez2021} report
a comparable case study for Otomi.  Notably, Moeller and Hulden's models
\emph{consume} a POS tier rather than evaluate it: word-level POS tags enter
their CRF as features (``and, of course, POS tags taken from the data,''
\citealp[p.~88]{moelleretal2018}), on
the explicit assumption that documentation data will come with POS.  Their
POS-informed CRF outperforms their no-POS seq2seq by 13 points, but the two
differ in architecture, so the tier's own contribution cannot be isolated;
our ablation turns exactly this assumed ingredient into a measured design
variable.  (Suggestively, their hardest labels are case morphemes such as
\gascore{gen} and \gascore{erg} --- precisely where our POS gain
concentrates; Appendix~\ref{app:perlabel}.)  Later work leverages the
\emph{translation} line \citep{mcmillanmajor2020,zhaoetal2020,yangetal2024}:
\citet{mcmillanmajor2020} generates the gloss line from the segmented source
phrase plus its English translation, \citet{zhaoetal2020} feed translations
to a multi-source transformer, and \citet{yangetal2024} quantify the tier's
effect: embedding LLM-encoded translations into a hard-attentional glosser
gains $+4.0$ points on average over the previous state of the art on the
SIGMORPHON 2023 data, and $+9.8$ points over the translation-free baseline
when training on as few as 100 sentences --- the same band as our POS-tier
gain ($+4.4$ with full data, $+11.6$ in the low-data regime), for a tier
that trilinear practice already produces.  A POS tier, by contrast, is far cheaper than
free translation (a closed inventory) yet is the tier standard practice
does not write down.  This strand culminated in the SIGMORPHON 2023
shared task on interlinear glossing \citep{ginnetal2023} and its systems
\citep[e.g.][]{girrbach2023}.  Most recently, massively
multilingual pretraining \citep{ginnetal2024} and LLM-based glossing
\citep{ginnetal2024b,elsneretal2025} have been explored --- the latter with
an interactive, linguist-assisting orientation close in spirit to ours
\citep{elsneretal2025} --- and \citet{aycocketal2025} probe gloss
prediction as one test of what LLMs learn about a low-resource language
from a grammar book.  This line of work
asks ``how accurate can glossing get given a corpus''; our question is the
converse and, to our knowledge, largely unaddressed: \emph{given a fixed,
small annotation budget, how should the annotation itself be designed ---
which tiers, how much?}  Our controlled comparison of annotation richness
(POS tier) and quantity, with seed-level significance testing and
error-propagation analysis, is complementary to all of the above.

\paragraph{Ryukyuan and Japanese dialects.}
To our knowledge, the only prior work on IGT automation for Japonic varieties
is our own proof-of-concept for Irabu \citep{shimojiryukyu49}, which
introduced the quality-dependent scenario, a joint segmentation--POS--gloss
prototype, and a dictionary on/off comparison, but no controlled ablation,
no multi-seed significance testing, and no analysis of error propagation or
annotation budgets.  Trained on a 30-minute subset of the present corpus
(289 sentences), that system reached 97.2\% word-level segmentation exact
match, 84.7\% gloss-token accuracy (dictionary-assisted, lexical and
grammatical glosses pooled), and 75.4\% POS accuracy on a 30-sentence
held-out set.  The numbers are not directly comparable to ours --- the data
volume, test sets, metrics, and task decomposition all differ --- but the
qualitative picture is consistent: under the present paper's stricter
evaluation (405 grammatical-gloss tokens, five seeds), the redesigned
pipeline improves POS tagging (0.75 $\rightarrow$ 0.88) and reaches
0.89--0.93 on grammatical glosses, superseding the prototype on all fronts.
Beyond IGT, NLP interest in endangered Japonic varieties is now emerging on
the \emph{translation} side: \citet{miyagawa2026} machine-translates the
Yoron dialect (Amami Islands) into Japanese with retrieval-augmented LLM
generation over roughly 9{,}000 phrase-aligned pairs drawn from a published
folk-tale collection.  That line exploits the phrase-translated (non-IGT)
resources that Japanese dialectology has produced in relative abundance, and
large proprietary LLMs; ours targets the annotation tiers that only IGT
provides --- segmentation, POS, and gloss --- with small supervised models.
The two are complementary halves of the same documentary bottleneck.

\paragraph{The documentation bottleneck.}
The programme of documentary linguistics --- compiling annotated corpora of
primary data for endangered languages --- was articulated by
\citet{himmelmann1998}; \citet{seifart2018} review its first twenty-five
years and identify the transcription bottleneck as a central obstacle.  On
the audio side of that bottleneck,
\citet{michaud2018} integrated automatic \emph{transcription} into the Na
documentation workflow; our work addresses the complementary
\emph{annotation} side (segmentation, POS, gloss), and the two lines compose
naturally into a single pipeline.  CRF-based morphological segmentation in
low-resource settings was established by \citet{ruokolainenetal2013}.
In Japan the bottleneck is also institutional: \citet{hidaka2026} surveys
ongoing collaborative efforts to digitize, align, and native-check legacy
dialect recordings --- most notably the 1977--1985 national emergency survey
(roughly ten hours of discourse at each of 228 sites) --- four decades after
their collection; we return to this in \S\ref{sec:conclusion}.

\section{Data and Pipeline}
\label{sec:data}

\subsection{Corpus}
Our corpus consists of 774 transcribed utterance units of Irabu Ryukyuan
(6{,}412 morpheme tokens), drawn from four recordings totalling 58 minutes
26 seconds of connected discourse.  The materials are entirely original
fieldwork by the author, recorded with five speakers of the Nagahama
district of Irabu island between 2005 and 2025; the genres are monologue ---
mostly legends and folktales --- and everyday conversation.  Transcription
and annotation were likewise carried out by the author: the corpus is fully
annotated with morpheme
boundaries, POS tags (32 tags, of which 31 occur in the training split;
full inventory in Appendix~\ref{app:tagset}), and
interlinear morphemic glosses
\citep{lehmann2004}; annotation was carried out in ELAN following the
workflow described in \citet{shimojiryukyu49}, whose corpus
(roughly 30 minutes) forms a subset of the present one.  The corpus also
carries a free-translation tier, but --- as stated in \S\ref{sec:intro} ---
no component of the pipeline uses it; all supervision comes from the
transcription, POS, and gloss tiers.
We distinguish \emph{grammatical glosses} (a closed set of category labels
such as \gascore{nom}, \gascore{gen}, \gascore{foc}; 115 types in training)
from \emph{lexical glosses} (open-class Japanese translations).  This paper's
quantitative target is the former; lexical positions are mapped to a
placeholder \gascore{lex} so sentence context is preserved.  Data are split
at utterance level into train/dev/test (620/77/77 utterances;
Table~\ref{tab:data}).  The gold corpus (\texttt{final.json}), all pipeline
and experiment code, and the complete per-sentence test outputs
(\texttt{test\_report.txt}) are available in the repository cited in
\S\ref{sec:intro} (\url{https://github.com/MichinoriShimoji/ML_autogloss}).

\begin{table}[t]
\centering
\small
\begin{tabular}{lrrrr}
\toprule
Split & Utt. & Tokens & Gram.\ tok. & Label types \\
\midrule
Train & 620 & 5{,}075 & 3{,}145 & 115 \\
Dev   &  77 &   675 &   432 &  76 \\
Test  &  77 &   662 &   405 &  64 \\
\bottomrule
\end{tabular}
\caption{Data splits.  The training set corresponds to roughly 47 minutes of
discourse and contains 836 morpheme types.}
\label{tab:data}
\end{table}

\subsection{Pipeline Components}
\label{sec:pipeline}

\begin{figure*}[t]
\centering
\resizebox{0.62\textwidth}{!}{%
\begin{tikzpicture}[>=stealth, font=\small,
  comp/.style={draw, rounded corners=3pt, inner sep=5pt, align=center, fill=gray!8},
  dat/.style={align=center, font=\small\itshape},
  sc/.style={font=\scriptsize, gray, align=center}]
  \node[dat]  (in)   at (0,0)    {word-segmented transcription\quad{\footnotesize\upshape jaanu accagaman c\textbari ffiutui}};
  \node[comp] (seg)  at (0,-1.4) {\textbf{1 Segmenter} --- char BiLSTM--CRF\\{\scriptsize\color{gray}span-F1 0.907 (vs.\ 0.726 longest-match)}};
  \node[dat]  (mor)  at (0,-2.8) {morphemes\quad{\footnotesize\upshape jaa nu acca gama n c\textbari ff i utui}};
  \node[comp] (pos)  at (0,-4.2) {\textbf{2 POS tagger} --- BiLSTM--CRF\\{\scriptsize\color{gray}accuracy 0.881}};
  \node[dat]  (post) at (0,-5.6) {POS tags\quad{\footnotesize\upshape\textsc{n cp n nafx cp v vafx vafx}}};
  \node[comp] (gls)  at (0,-7.2) {\textbf{3 Gloss labeler}\\BiLSTM--CRF over morpheme$\,\oplus\,$POS\\{\scriptsize\color{gray}gram-acc 0.93 (gold POS) / 0.89 (pipeline)}};
  \node[dat]  (out)  at (0,-10.6) {quadrilinear IGT\\[2pt]
    {\footnotesize\upshape
    \begin{tabular}{@{}lll@{}}
      \itshape jaa=nu & \itshape acca-gama=n & \itshape c\textbari ff-i-utui\\
      \textsc{n}=\textsc{cp} & \textsc{n}-\textsc{nafx}=\textsc{cp} & \textsc{v}-\textsc{vafx}-\textsc{vafx}\\
      house=\textsc{gen} & side-\textsc{dim}=\textsc{dat} & make-\textsc{thm}-\textsc{crcm}\\
      \multicolumn{3}{@{}l@{}}{`(they) were making (it) by the house'}\\
    \end{tabular}}};
  \node[comp] (bnd) at (5.8,-5.6) {\textbf{Boundary types}\\rule: next POS $\in$ particle\\classes $\Rightarrow$ \emph{=}, else \emph{-}\\{\scriptsize\color{gray}97.6\% (gold POS)}};
  \node[comp] (lex) at (5.8,-8.7) {\textbf{4 Lexical glosses}\\lexicon copy + reranking;\\dictionary fallback\\{\scriptsize\color{gray}exact match 0.69}};
  \draw[->] (in) -- (seg);
  \draw[->] (seg) -- (mor);
  \draw[->] (mor) -- (pos);
  \draw[->] (pos) -- (post);
  \draw[->] (post) -- (gls);
  \draw[->] (gls) -- (out);
  \draw[->] (post.east) -- (bnd.west);
  \draw[->] (gls.south east) -- node[sc, below, sloped] {\gascore{lex} positions} (lex.west);
  \draw[->] (bnd.east) .. controls (8.6,-5.6) and (8.6,-10.6) .. (out.east);
  \draw[->] (lex.south) .. controls (5.8,-10.2) and (4.6,-10.6) .. (out.east);
\end{tikzpicture}}
\caption{The annotation pipeline.  Numbered boxes are trained BiLSTM--CRF
components (test scores below each); the boundary-type module is a rule over
the predicted POS of the attaching morpheme (\S\ref{sec:pipeline}).  Gloss
labeling consumes both the morphemes and their POS tags --- the dependency
whose value and cost this paper quantifies.  The lexical-gloss module is
peripheral to our claims.  The worked example is a real fragment of the
corpus (`(they) were making (it) by the house', from a wartime narrative),
and every tier shown is the pipeline's actual output on that fragment, which
here coincides with the gold annotation.}
\label{fig:pipeline}
\end{figure*}
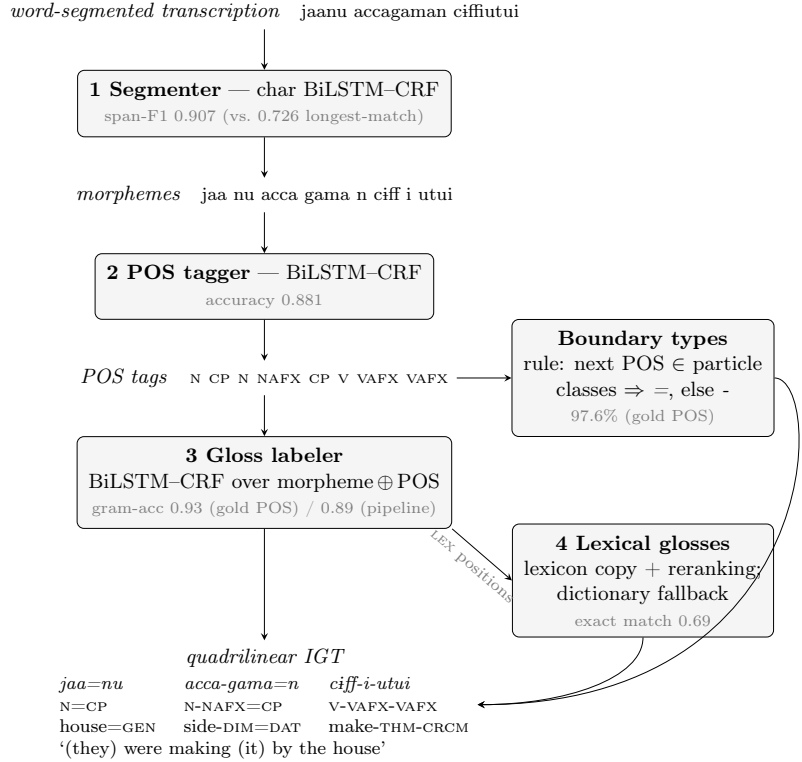

The pipeline (Figure~\ref{fig:pipeline}) mirrors the interlinearization
workflow: (1)~a character-level
BiLSTM--CRF segments words into morphemes (test span-F1 0.907, vs.\ 0.726 for
a longest-match dictionary baseline; $p<10^{-4}$, paired
randomization/bootstrap); (2)~a BiLSTM--CRF POS tagger over morphemes (test
accuracy 0.881); (3)~the glossing model described next.  A lexical-gloss
module (training-lexicon copy with contextual reranking plus a fallback to a
digitized dictionary of a neighboring variety, \citealp{tomihama2013})
reaches 0.69 exact match but is peripheral here.\footnote{The
dictionary (15{,}132 entries) was digitized from the printed original with
the National Diet Library's NDLOCR-Lite engine, followed by custom OCR
post-correction and headword-structure extraction; residual OCR noise is one
reason the dictionary serves as a fallback rather than an oracle.}

\paragraph{Boundary types come from POS.}
IGT distinguishes affixal boundaries (\emph{-}) from clitic boundaries
(\emph{=}), so a segmenter must in principle deliver typed boundaries.  Our
design factors this apart: the segmenter predicts boundary \emph{positions}
only, and the \emph{type} is derived from the POS of the following morpheme
--- particle classes (case, focus, final, conjunctive, etc.) attach with
\emph{=}, affix classes with \emph{-}.  In our corpus this majority rule
reconstructs the type of 97.6\% of all 3{,}613 word-internal boundaries
(and, fit on the training split alone, scores 97.8\% on the 368 held-out
test boundaries): the affix/clitic distinction is, to a close approximation,
a corollary of lexical category.  The alternative --- a segmenter trained to
predict typed boundaries jointly (eight labels instead of four) --- is
empirically just as good at segmenting: across 5 seeds, position span-F1 is
$0.901\pm0.013$ (typed) versus $0.900\pm0.009$ (plain), and type accuracy at
correctly detected positions averages 97.8\%.\footnote{An earlier draft
reported a 1.3-point position cost for the typed segmenter; a 5-seed
replication (released with our code) shows that difference was single-seed
noise.}  Which route should deliver the type therefore depends on the
quality of the POS the rule consumes, and we measured both (test set, 368
boundaries; experiment released with our code).  On \emph{gold} POS the
rule ties the typed segmenter (97.8\%).  On \emph{predicted} POS it falls
to 95.4\% --- whether segmentation is gold or predicted --- while the typed
segmenter, which never consults POS, keeps $97.8\pm0.4$\% at the positions
it detects: a $+2.4$-point advantage, consistent in direction in all 5
seeds (pooled discordant boundaries 60:17; exact McNemar significant in 2
of 5 seeds).  For a \emph{fully automatic} pipeline, then, typed
segmentation is the better engineering choice.  Our deployed interactive
tool still derives types from POS, for a workflow reason: keeping one
source of truth means a POS correction automatically corrects the attached
boundary type, and once the annotator has reviewed the POS tier --- which
the Stage-2 workflow requires anyway --- the rule operates on gold-quality
tags and recovers its 97.8\% ceiling.  This is a first instance of the
paper's recurring theme --- \emph{the POS tier quietly carries other tiers
of IGT with it} --- and, in the same breath, of its qualification: it
carries them at gold quality, and tagger errors tax the ride
(\S\ref{sec:propagation} quantifies the same pattern for glossing).
\label{sec:experiments}

We ask whether a POS tier --- quadrilinear rather than trilinear annotation
--- improves automatic grammatical glossing.  We separate two questions:
(A)~does POS information have intrinsic value for gloss prediction, and
(B)~can that value be realized in a fully automatic pipeline, where POS tags
themselves must be predicted?

\subsection{Glossing Model and Conditions}
\label{sec:conditions}

The glossing model is a BiLSTM--CRF over morphemes: each morpheme is
represented by a 64-dimensional embedding, optionally concatenated with a
16-dimensional POS embedding, fed to a single-layer BiLSTM (hidden size 128)
with a CRF output layer.  Note where each information source acts: POS is
visible \emph{only to the BiLSTM}, as an input feature; the CRF operates
purely over the output gloss labels, scoring adjacent label pairs, and never
sees POS.  All other factors --- data splits, architecture,
optimizer (Adam, lr $10^{-3}$), batch size (16), training length (40 epochs,
best-dev selection) --- are held constant; the \emph{only} manipulated factor
is the POS input.  Three conditions:

\begin{itemize}
\item \textbf{Gold POS} (oracle): human POS tags at training and test time.
  Measures the \emph{intrinsic value} of POS (question~A).  Not available in
  deployment, but operative in computer-assisted annotation, where a human
  supplies POS before the model proposes glosses.
\item \textbf{Predicted POS} (pipeline): test-time tags from the POS tagger
  of \S\ref{sec:data} (accuracy 0.881).  Measures \emph{realizable value}
  (question~B), including error propagation.  The information flow is
  identical to the Gold-POS condition: the tagger first Viterbi-decodes the
  complete POS sequence for the sentence, and the glossing model then
  receives those tags exactly as it would receive gold ones --- the two
  conditions differ in input quality, not in architecture; nothing is
  predicted jointly.
\item \textbf{No POS} (control): the POS embedding is ablated.
\end{itemize}

Both models are trained from scratch with five random seeds (mean $\pm$ SD
reported).  Within each seed, Gold-POS and No-POS systems are compared on the
same 405 test tokens with an exact McNemar test.

\subsection{Main Results (RQ1: Richness)}

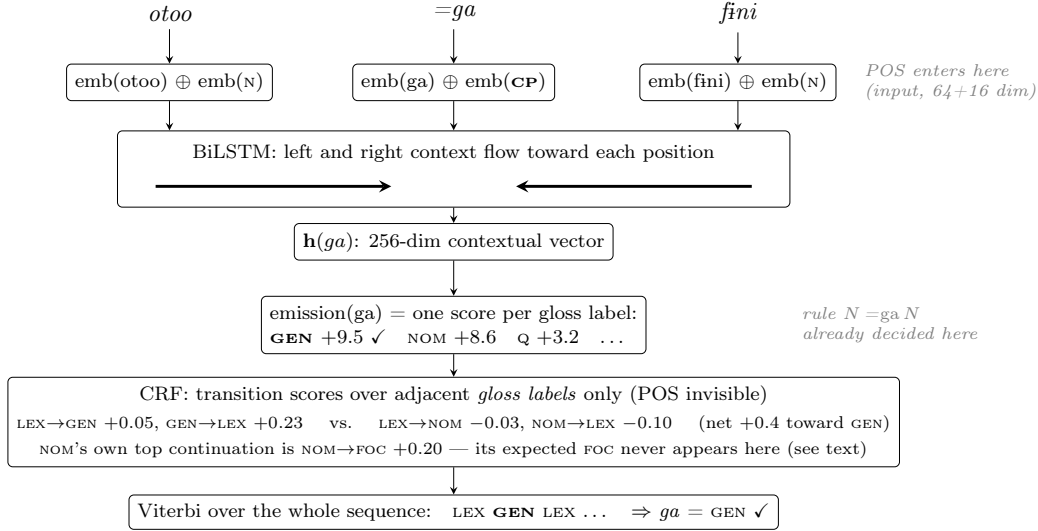
\begin{figure*}[t]
\centering
\resizebox{0.78\textwidth}{!}{%
\begin{tikzpicture}[>=stealth, font=\small,
  tok/.style={draw, rounded corners=2pt, inner sep=3.5pt, align=center, font=\footnotesize},
  note/.style={font=\scriptsize\itshape, gray, align=left},
  stage/.style={font=\scriptsize\sffamily, gray}]
  \node[font=\normalsize\itshape] (t1) at (0,6.1)   {otoo};
  \node[font=\normalsize\itshape] (t2) at (4.1,6.1) {=ga};
  \node[font=\normalsize\itshape] (t3) at (8.2,6.1) {f\textbari ni};
  \node[tok] (i1) at (0,5.1)   {emb(otoo) $\oplus$ emb(\textsc{n})};
  \node[tok] (i2) at (4.1,5.1) {emb(ga) $\oplus$ emb(\textbf{\textsc{cp}})};
  \node[tok] (i3) at (8.2,5.1) {emb(f\textbari ni) $\oplus$ emb(\textsc{n})};
  \node[note, anchor=west] at (9.9,5.1) {POS enters here\\(input, 64+16 dim)};
  \draw[->] (t1) -- (i1); \draw[->] (t2) -- (i2); \draw[->] (t3) -- (i3);
  \node[tok, minimum width=9.7cm, minimum height=1.1cm] (lstm) at (4.1,3.85) {};
  \node[font=\footnotesize] at (4.1,4.08) {BiLSTM: left and right context flow toward each position};
  \draw[->, very thick] (-0.2,3.6) -- (3.2,3.6);
  \draw[->, very thick] (8.4,3.6) -- (5.0,3.6);
  \draw[->] (i1) -- (i1 |- lstm.north);
  \draw[->] (i2) -- (i2 |- lstm.north);
  \draw[->] (i3) -- (i3 |- lstm.north);
  \node[tok] (h) at (4.1,2.8) {$\mathbf{h}(ga)$: 256-dim contextual vector};
  \draw[->] (lstm) -- (h);
  \node[tok, align=left] (em) at (4.1,1.6)
    {emission(ga) = one score per gloss label:\\
     \textbf{\gascore{gen} $+9.5$} \checkmark\quad \gascore{nom} $+8.6$\quad \gascore{q} $+3.2$\quad \ldots};
  \draw[->] (h) -- (em);
  \node[note, anchor=west] at (9.0,1.6) {rule $N$\,=\emph{ga}\,$N$\\already decided here};
  \node[tok, minimum width=9.7cm, align=center] (crf) at (4.1,0.2)
    {CRF: transition scores over adjacent \emph{gloss labels} only (POS invisible)\\[2pt]
     {\scriptsize \gascore{lex}$\rightarrow$\gascore{gen} $+0.05$, \gascore{gen}$\rightarrow$\gascore{lex} $+0.23$ \quad vs.\ \quad \gascore{lex}$\rightarrow$\gascore{nom} $-0.03$, \gascore{nom}$\rightarrow$\gascore{lex} $-0.10$ \quad (net $+0.4$ toward \gascore{gen})}\\[1pt]
     {\scriptsize \gascore{nom}'s own top continuation is \gascore{nom}$\rightarrow$\gascore{foc} $+0.20$ --- its expected \gascore{foc} never appears here (see text)}};
  \draw[->] (em) -- (crf);
  \node[tok] (out) at (4.1,-1.1)
    {Viterbi over the whole sequence:\quad \gascore{lex}\;\ \textbf{\gascore{gen}}\;\ \gascore{lex}\ \ldots\quad$\Rightarrow$ \emph{ga} = \gascore{gen} \checkmark};
  \draw[->] (crf) -- (out);
\end{tikzpicture}}
\caption{How the trained model glosses \emph{ga} in example (\ref{ex:gen}),
with actually measured scores.  POS participates only as an input feature:
the BiLSTM merges the flanking nouns' categories into \emph{ga}'s contextual
vector, and the resulting emission already ranks \gascore{gen} above
\gascore{nom} ($+9.5$ vs.\ $+8.6$; the emission-only argmax is already
\gascore{gen}).  The CRF sees no POS; it scores adjacent gloss labels, yet
its learned transitions also favor \gascore{gen} here (net $+0.4$) ---
syntax echoed in label-sequence statistics.  Contribution to the decision:
roughly $2/3$ emission, $1/3$ transitions.}
\label{fig:mechanism}
\end{figure*}

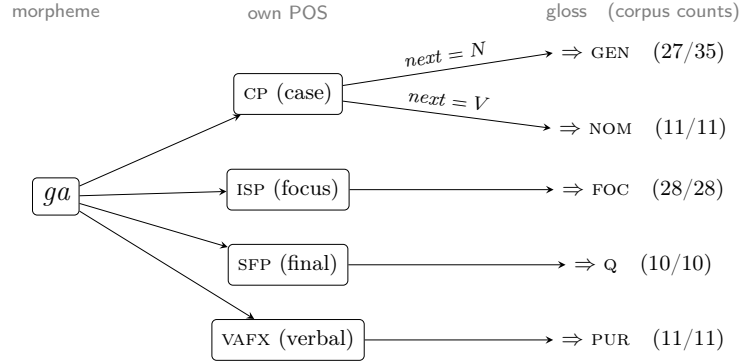
\begin{figure*}[t]
\centering
\resizebox{0.56\textwidth}{!}{%
\begin{tikzpicture}[>=stealth, font=\small,
  lab/.style={font=\footnotesize\itshape, midway, above, sloped},
  box/.style={draw, rounded corners=2pt, inner sep=4pt}]
  \node[box, font=\large\itshape] (ga) at (0,0) {ga};
  \node[box] (cp)   at (3.4, 1.5) {\textsc{cp} (case)};
  \node[box] (isp)  at (3.4, 0.1) {\textsc{isp} (focus)};
  \node[box] (sfp)  at (3.4,-1.0) {\textsc{sfp} (final)};
  \node[box] (vafx) at (3.4,-2.1) {\textsc{vafx} (verbal)};
  \node (gen) at (8.6, 2.1) {$\Rightarrow$ \gascore{gen}\quad (27/35)};
  \node (nom) at (8.6, 1.0) {$\Rightarrow$ \gascore{nom}\quad (11/11)};
  \node (foc) at (8.6, 0.1) {$\Rightarrow$ \gascore{foc}\quad (28/28)};
  \node (q)   at (8.6,-1.0) {$\Rightarrow$ \gascore{q}\quad (10/10)};
  \node (pur) at (8.6,-2.1) {$\Rightarrow$ \gascore{pur}\quad (11/11)};
  \draw[->] (ga) -- (cp);
  \draw[->] (ga) -- (isp);
  \draw[->] (ga) -- (sfp);
  \draw[->] (ga) -- (vafx);
  \draw[->] (cp) -- node[lab] {next $= N$} (gen.west);
  \draw[->] (cp) -- node[lab] {next $= V$} (nom.west);
  \draw[->] (isp)  -- (foc.west);
  \draw[->] (sfp)  -- (q.west);
  \draw[->] (vafx) -- (pur.west);
  \node[font=\footnotesize\sffamily, gray] at (0, 2.7) {morpheme};
  \node[font=\footnotesize\sffamily, gray] at (3.4, 2.7) {own POS};
  \node[font=\footnotesize\sffamily, gray] at (8.6, 2.7) {gloss\quad(corpus counts)};
\end{tikzpicture}}
\caption{How POS determines the gloss of the six-ways ambiguous clitic
\emph{ga} (counts over the full corpus).  Its own POS tag resolves three of
the four readings outright --- \textsc{isp}, \textsc{sfp}, and \textsc{vafx}
force \gascore{foc}, \gascore{q}, and \gascore{pur} without exception ---
and for the case-particle reading (\textsc{cp}), the category of the
following word decides: noun $\rightarrow$ \gascore{gen} (typical;
exceptions are nominal-predicate clauses), verb $\rightarrow$ \gascore{nom}
(without exception).  The remaining \textsc{cp} contexts (e.g.\ before focus
clitics or utterance-finally) are \gascore{nom} in 13 of 16 cases.}
\label{fig:ga}
\end{figure*}
\label{sec:results}

\begin{table}[t]
\centering
\begin{tabular}{lc}
\toprule
Condition & Gram.\ gloss accuracy \\
\midrule
Gold POS       & \textbf{0.937 $\pm$ 0.004} \\
Predicted POS  & 0.897 $\pm$ 0.006 \\
No POS         & 0.893 $\pm$ 0.008 \\
\midrule
$\Delta$(Gold $-$ No)      & $+4.35 \pm 0.73$ pts \\
$\Delta$(Predicted $-$ No) & $+0.40 \pm 0.77$ pts \\
\bottomrule
\end{tabular}
\caption{Grammatical gloss accuracy on test (405 tokens), mean $\pm$ SD over
5 seeds.}
\label{tab:main}
\end{table}

\begin{table}[t]
\centering
\resizebox{\columnwidth}{!}{%
\begin{tabular}{lccccc}
\toprule
Seed & Gold & Pred. & No POS & $\Delta$(Gold$-$No) & McNemar $p$ \\
\midrule
1 & 0.936 & 0.891 & 0.889 & $+4.7$ & $1.6\times10^{-4}$ \\
2 & 0.943 & 0.904 & 0.901 & $+4.2$ & $7.6\times10^{-5}$ \\
3 & 0.933 & 0.894 & 0.882 & $+5.2$ & $1.9\times10^{-5}$ \\
4 & 0.933 & 0.894 & 0.901 & $+3.2$ & $7.2\times10^{-3}$ \\
5 & 0.938 & 0.904 & 0.894 & $+4.4$ & $2.8\times10^{-4}$ \\
\bottomrule
\end{tabular}}
\caption{Per-seed results.  The Gold-POS advantage is positive and
individually significant ($p<0.01$) in every seed; discordant pairs
consistently favor the POS-informed model (22:3, 18:1, 23:2, 17:4, 21:3).}
\label{tab:seeds}
\end{table}

With gold POS, the POS-informed model outperforms the ablated model by
$+4.35\pm0.73$ points; the gain is positive and individually significant in
all five seeds (Tables~\ref{tab:main}--\ref{tab:seeds}; all $p<0.01$, hence
significant even under Bonferroni correction for the five tests).  \emph{POS
information is intrinsically valuable for grammatical glossing}
(question~A: yes).

\paragraph{Why POS helps: the case of \emph{ga}.}
The Irabu clitic \emph{ga} occurs 112 times in our corpus under six distinct
grammatical glosses, and its segmental form gives no clue.  Yet with POS in
hand, the ambiguity dissolves into rules a linguist would state on one line
(all counts are corpus totals):

\begin{itemize}
\item \textbf{If \emph{ga} is a case particle (\textsc{cp}), the next word
  decides:} $N$~\emph{=ga}~$N$ $\rightarrow$ \gascore{gen} (`X's Y'; 27/35),
  whereas $N$~\emph{=ga}~$V$ $\rightarrow$ \gascore{nom} (subject; 11/11).
\item \textbf{If \emph{ga}'s own POS is \textsc{isp}, the gloss is forced:}
  focus particle \emph{ga} is \gascore{foc} \emph{without exception}
  (28/28).  Likewise \textsc{sfp} $\rightarrow$ \gascore{q} (10/10) and
  \textsc{vafx} $\rightarrow$ \gascore{pur} (11/11).
\end{itemize}

Figure~\ref{fig:ga} schematizes this disambiguation, and
(\ref{ex:gen})--(\ref{ex:foc}) are real corpus utterances instantiating it:
in (\ref{ex:gen}) \emph{ga} sits between two nouns (\gascore{gen}), in
(\ref{ex:nom}) a verb follows (\gascore{nom}), and in (\ref{ex:foc})
\emph{ga} itself is a focus particle (\textsc{isp}), which forces
\gascore{foc}.  Note that (\ref{ex:nom}) and (\ref{ex:foc}) are string-wise
near-parallel (\emph{\ldots=ga \ldots-tar=gagara=mmja}); only the POS tier
tells them apart.

{\small
\begin{exe}
\ex \label{ex:gen}
\glll otoo {\textbf{=ga}} f\textbari ni =nu =du f\textbari\textbari{} =ti \\
     {\textsc{n}} {\textbf{\textsc{cp}}} {\textsc{n}} {\textsc{cp}} {\textsc{isp}} {\textsc{v}} {\textsc{cjp}} \\
     father {\textbf{\gascore{gen}}} boat {\gascore{nom}} {\gascore{foc}} come {\gascore{quot}} \\
\glt `(They said) father's boat is coming.'
\ex \label{ex:nom}
\glll taa {\textbf{=ga}} tur -tar =gagara =mmja \\
     {\textsc{prn}} {\textbf{\textsc{cp}}} {\textsc{v}} {\textsc{vafx}} {\textsc{mp}} {\textsc{intj}} \\
     who {\textbf{\gascore{nom}}} take {\gascore{pst}} {\gascore{dub}} {\gascore{dsc}} \\
\glt `Who took it, I wonder.'
\ex \label{ex:foc}
\glll unu nausi {\textbf{=ga}} maf\textbari{} -tar =gagara =mmja \\
     {\textsc{intj}} {\textsc{adv}} {\textbf{\textsc{isp}}} {\textsc{v}} {\textsc{vafx}} {\textsc{mp}} {\textsc{intj}} \\
     {\gascore{fil}} how {\textbf{\gascore{foc}}} wind {\gascore{pst}} {\gascore{dub}} {\gascore{dsc}} \\
\glt `How did it wind up, I wonder.'
\end{exe}}

These category-level rules are exactly what the POS-informed model can
exploit and what a morpheme-only model cannot see: on the test set, the
ablated model misglossed \emph{ga} as \gascore{nom} in both (\ref{ex:gen})
and (\ref{ex:foc}) --- the actual errors we observed --- and resolved only 7
of 10 grammatical \emph{ga} tokens overall, versus 10/10 with POS.\footnote{How
does the \emph{tagger} itself resolve \emph{ga}, having no POS tier of its
own?  It must reconstruct category information on the fly, from two implicit
sources: morpheme embeddings, which POS supervision shapes into soft
category proxies (noun-like morphemes cluster together), and the CRF's joint
decoding, in which the hypothesized tags of \emph{ga} and its neighbors
constrain one another as a single sequence.  The reconstruction is
frequency-weighted: on the ten grammatical \emph{ga} tokens in test the
tagger scores 7/10, succeeding after nominal hosts ($\rightarrow$
\textsc{cp}) and after another case clitic --- where \textsc{isp} is forced
because a case clitic follows a case clitic in only 2 of 512 training
bigrams --- and defaulting to majority \textsc{cp} in the three rare left
contexts, including example~(\ref{ex:foc}).  Reconstructed category is also
what collapses on unseen morphemes (tagger accuracy 0.90 known vs.\ 0.70
unknown).  An explicit POS tier hands the model this category directly
instead of asking it to be reconstructed from lexical identity --- which is
why its value grows precisely as data shrink (\S\ref{sec:datasize}).}
Appendix~\ref{app:perlabel} confirms the pattern corpus-wide: the largest
per-label gains fall exactly on the case and focus/topic clitics
(Table~\ref{tab:glosslabel}).

\paragraph{Where in the model the rule lives.}
It is worth being precise about \emph{where} the $N$~\emph{=ga}~$N$ rule is
computed, because the answer illustrates the transparency we appealed to in
\S\ref{sec:intro}.  Figure~\ref{fig:mechanism} traces example
(\ref{ex:gen}) through the trained model, with actually measured scores.
POS enters at the very bottom, as an input embedding concatenated to each
morpheme embedding.  The BiLSTM then folds the left context (\emph{otoo},
\textsc{n}) and the right context (\emph{f\textbari ni}, \textsc{n}) into
\emph{ga}'s contextual vector, and a linear layer converts that vector into
\emph{emission scores} --- one score per gloss label.  The categorical rule
is already decided at this point: \gascore{gen} scores $+9.5$ against
\gascore{nom}'s $+8.6$, so even before the CRF, the emission argmax is
\gascore{gen}.  The CRF, which never sees POS, then adds transition scores
between adjacent \emph{output labels} and Viterbi-decodes the globally best
gloss sequence.  Interestingly, its transitions push in the same direction
(net $+0.4$), and the reason is instructive.  At the label level the two
competitors look identical here --- \gascore{lex}~\gascore{gen}~\gascore{lex}
vs.\ \gascore{lex}~\gascore{nom}~\gascore{lex} --- so one might expect the
CRF to be indifferent.  It is not, because the two labels have sharply
different \emph{continuation profiles} in the training data: \gascore{gen},
being adnominal, is followed by a lexical gloss 95\% of the time (118/124),
whereas \gascore{nom} is followed by a lexical gloss only 64\% of the time
(43/67), with much of the remaining mass on \gascore{foc} (18\%) ---
reflecting the Miyako focus construction in which a nominative subject hosts
the focus clitic \emph{=du}, visible in example~(\ref{ex:gen}) itself
(\emph{f\textbari ni=nu=du}).  The learned transitions mirror this exactly:
\gascore{gen}$\rightarrow$\gascore{lex} ($+0.23$) is \gascore{gen}'s
highest-scoring continuation, while \gascore{nom}'s is
\gascore{nom}$\rightarrow$\gascore{foc} ($+0.20$), leaving
\gascore{nom}$\rightarrow$\gascore{lex} at $-0.10$.  In a
\gascore{lex}\,\_\,\gascore{lex} frame, then, the CRF favors \gascore{gen}
not because it sees the syntax, but because \gascore{nom}'s expected
continuation (\gascore{foc}) fails to appear --- a distributional echo of
the syntax in the label space, learned without ever seeing a POS tag.  The
decision thus decomposes into interpretable parts: roughly two thirds
context-driven emission (the POS-informed BiLSTM) and one third
label-sequence prior (the CRF).

With \emph{predicted} POS the gain shrinks to $+0.40\pm0.77$ points ---
smaller than its seed variance, one seed negative, and non-significant by an
exact McNemar test in every seed (smallest $p=0.44$).  The current pipeline
does not realize the value of POS (question~B: not yet).
\S\ref{sec:propagation} locates the loss.

\subsection{Analysis: Where the 4 Points Are Lost}
\label{sec:propagation}

\begin{table}[t]
\centering
\resizebox{\columnwidth}{!}{%
\begin{tabular}{lccc}
\toprule
 & \multicolumn{3}{c}{Gloss accuracy at these positions} \\
\cmidrule(lr){2-4}
Position type & Gold POS & Predicted POS & No POS \\
\midrule
Tagger correct (371 tok., 91.6\%) & 0.963 & 0.960 & 0.930 \\
Tagger error (34 tok., 8.4\%)     & 0.653 & \textbf{0.218} & \textbf{0.494} \\
\bottomrule
\end{tabular}}
\caption{Seed-averaged gloss accuracy, split by whether the tagger predicted
the token's own POS correctly.  At mistagged positions an incorrect POS
feature is \emph{worse than no POS at all}.}
\label{tab:positions}
\end{table}

Gold and predicted POS differ only where the tagger errs, so the 4-point gap
must originate there (the released test report shows the propagation
sentence by sentence).  To see how, we split the 405 test tokens into two
groups --- the 371 (91.6\%) where the tagger predicted the token's own POS
correctly, and the 34 (8.4\%) where it did not --- and ask how each model
fares in each group (Table~\ref{tab:positions}, visualized in
Figure~\ref{fig:budget}).  Three findings follow.

\paragraph{(a) The damage is localized.}
Where the tagger is right, nothing is lost: the POS-informed model performs
as well with predicted POS as with gold (0.960 vs.\ 0.963).  Essentially all
of the damage sits in the small mistagged group: there, switching from gold
to predicted POS flips 44.1\% of the glosses from correct to incorrect,
against a negligible 0.4\% elsewhere.

\paragraph{(b) At those positions, a wrong POS is worse than no POS.}
This is the key finding (Figure~\ref{fig:budget}, top).  One might expect a
wrong POS to be merely useless.  It is not: at mistagged positions the model
that receives the (wrong) POS scores 0.218, far \emph{below} the 0.494 of
the model that receives no POS at all.  A wrong tag is \emph{misleading
evidence} --- the model, trained exclusively on gold POS, has never seen an
incorrect tag and follows it off a cliff (a train/test mismatch akin to
exposure bias).

\paragraph{(c) The two effects balance almost exactly.}
The observed pipeline gain is now just bookkeeping (Figure~\ref{fig:budget},
bottom): on the 91.6\% of well-tagged tokens, POS earns $3.0$ points
per token ($0.9596$ vs.\ $0.9299$); on the 8.4\% of mistagged tokens it
\emph{loses} $27.6$ points per token ($0.2176$ vs.\ $0.4941$).  Weighting by
group size,
\begin{equation}
\begin{split}
&\underbrace{(0.9596-0.9299)\times0.916}_{+2.72\ \text{pts (gain)}}
+
\underbrace{(0.2176-0.4941)\times0.084}_{-2.32\ \text{pts (poison)}}\\
&\qquad = +0.40\ \text{pts},
\end{split}
\label{eq:budget}
\end{equation}
which reproduces the observed $+0.40$ exactly.  In words: \emph{each tagger
error is roughly nine times as damaging as each correct tag is helpful, so
8\% poison cancels 92\% gain.}  The pipeline gain vanishes not because POS
lacks value, but because of where and how the errors strike.

\begin{figure*}[t]
\centering
\resizebox{0.58\textwidth}{!}{%
\begin{tikzpicture}[font=\small]
  \node[font=\small\sffamily] at (4.6,3.35)
    {Gloss accuracy at the 34 \emph{mistagged} positions};
  \node[anchor=east] at (2.45,2.7) {Gold POS};
  \fill[teal!60]  (2.6,2.52) rectangle (6.52,2.88);   %
  \node[anchor=west] at (6.6,2.7) {0.653};
  \node[anchor=east] at (2.45,2.1) {No POS};
  \fill[gray!55]  (2.6,1.92) rectangle (5.56,2.28);   %
  \node[anchor=west] at (5.64,2.1) {0.494};
  \node[anchor=east] at (2.45,1.5) {Wrong POS (pred.)};
  \fill[red!60]   (2.6,1.32) rectangle (3.91,1.68);   %
  \node[anchor=west] at (3.99,1.5) {\textbf{0.218} --- worse than no POS};
  \node[font=\small\sffamily] at (4.6,0.55)
    {The budget: why the pipeline gain is only $+0.40$};
  \node[draw=teal, rounded corners=3pt, inner sep=5pt, align=center]
    (gain) at (1.9,-0.65)
    {\textbf{gain} on 91.6\%\\ of tokens\\ $+2.97 \times 0.916$\\ $= +2.72$~pts};
  \node[font=\Large] at (4.05,-0.65) {$-$};
  \node[draw=red!70, rounded corners=3pt, inner sep=5pt, align=center]
    (loss) at (6.0,-0.65)
    {\textbf{poison} on 8.4\%\\ of tokens\\ $-27.6 \times 0.084$\\ $= -2.32$~pts};
  \node[font=\Large] at (8.05,-0.65) {$=$};
  \node[draw, thick, rounded corners=3pt, inner sep=6pt, align=center]
    at (9.9,-0.65) {$+0.40$~pts\\ {\scriptsize(= observed)}};
\end{tikzpicture}}
\caption{Why tagger errors cancel the POS gain.  Top: at the 34 positions
where the tagger mistags the token itself, receiving the wrong POS (0.218)
is far worse than receiving none (0.494) --- an incorrect feature actively
misleads the model.  Bottom: the small gain earned on the many well-tagged
tokens is almost exactly cancelled by the deep loss on the few mistagged
ones (Eq.~\ref{eq:budget}).}
\label{fig:budget}
\end{figure*}

\subsection{Break-Even Analysis: How Good Must the Tagger Be?}
\label{sec:breakeven}

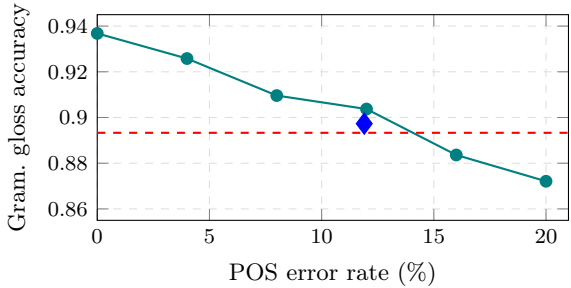
\begin{figure}[t]
\centering
\begin{tikzpicture}
\begin{axis}[
  width=0.92\linewidth, height=4.3cm,
  tick label style={font=\footnotesize},
  label style={font=\small},
  xlabel={POS error rate (\%)}, ylabel={Gram.\ gloss accuracy},
  xmin=0, xmax=21, ymin=0.855, ymax=0.945,
  grid=major, grid style={dashed,gray!25},
]
\addplot[color=teal, mark=*, thick] coordinates {
  (0,0.9368) (4,0.9258) (8,0.9096) (12,0.9037) (16,0.8836) (20,0.8721)
};
\addplot[color=red, dashed, thick, domain=0:21] {0.8933};
\addplot[color=blue, mark=diamond*, only marks, mark size=4pt] coordinates {
  (11.9,0.8973)
};
\end{axis}
\end{tikzpicture}
\caption{Gloss accuracy vs.\ POS error rate.  Teal curve: POS-informed
model on noise-injected gold POS (noise sampled from the real tagger's
confusion distribution, estimated on dev; means over 5 seeds $\times$ 3
draws).  Red dashed line: No-POS baseline (0.8933).  Blue diamond: the real
tagger (measured).  Simulated break-even: 14.1\% error (85.9\% accuracy).
The real tagger (11.9\% error) falls \emph{below} the curve because real
errors concentrate on intrinsically difficult tokens.}
\label{fig:breakeven}
\end{figure}

Injecting controlled noise into gold POS --- substitution errors sampled from
the real tagger's confusion distribution --- traces how tagger quality
translates into gloss quality (Figure~\ref{fig:breakeven}).  The POS-informed
model beats the baseline by $+3.2$ points at a 4\% error rate and $+1.6$
points at 8\%, crossing it at 14.1\%.  Two caveats sharpen the picture.
First, the real tagger (11.9\% error) yields only $+0.40$, below the
simulated curve at the same rate: real errors cluster on hard tokens, so the
effective break-even for a real tagger is nearer 88\% accuracy --- almost
exactly where our tagger stands (88.1\%).  Second, the curve sets a concrete
target: a tagger at 92--96\% accuracy converts the latent value into
$+1.6$ to $+3.2$ practical points.

\subsection{Interaction with Data Size (RQ2): POS as a Data Multiplier}
\label{sec:datasize}

If POS acts as a coarse generalization class, its value should be greatest
when training data are scarce --- the situation at the start of a
documentation project.  We train both conditions on nested subsets of the
training data (78, 155, 310, 620 utterances $\approx$ 6, 12, 23, 47 minutes
of discourse), rebuilding all vocabularies from each subset; five seeds per
cell, gold POS at test.

\begin{table}[t]
\centering
\small
\begin{tabular}{rrccc}
\toprule
Train.\ utt. & $\approx$min & Gold POS & No POS & $\Delta$ (pts) \\
\midrule
 78 &  6 & 0.765 & 0.671 & $+9.5$ \\
155 & 12 & 0.839 & 0.723 & $+11.6$ \\
310 & 23 & 0.890 & 0.822 & $+6.8$ \\
620 & 47 & 0.933 & 0.888 & $+4.5$ \\
\bottomrule
\end{tabular}
\caption{Learning curves (test accuracy, 5-seed means).  The POS gain is
largest in the low-data regime: 2.6$\times$ larger at 12 minutes than at 47.}
\label{tab:datasize}
\end{table}

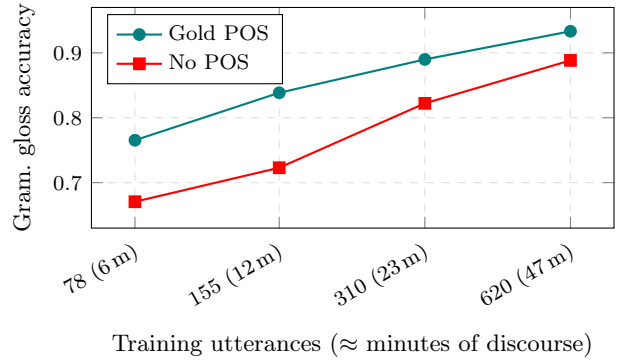
\begin{figure}[t]
\centering
\begin{tikzpicture}
\begin{axis}[
  width=\linewidth, height=4.5cm,
  tick label style={font=\footnotesize},
  label style={font=\small},
  legend style={font=\footnotesize},
  xlabel={Training utterances ($\approx$ minutes of discourse)},
  ylabel={Gram.\ gloss accuracy},
  xmode=log, log basis x=2,
  xtick={78,155,310,620},
  xticklabels={78 (6\,m),155 (12\,m),310 (23\,m),620 (47\,m)},
  x tick label style={font=\footnotesize, rotate=30, anchor=north east},
  ymin=0.63, ymax=0.97,
  legend pos=north west, legend cell align=left,
  grid=major, grid style={dashed,gray!25},
]
\addplot[color=teal, mark=*, thick] coordinates {
  (78,0.7654) (155,0.8385) (310,0.8899) (620,0.9333)
};
\addlegendentry{Gold POS}
\addplot[color=red, mark=square*, thick] coordinates {
  (78,0.6706) (155,0.7230) (310,0.8222) (620,0.8884)
};
\addlegendentry{No POS}
\end{axis}
\end{tikzpicture}
\caption{Learning curves.  The POS-informed model trained on 310 utterances
(0.890) already exceeds the ablated model trained on all 620 (0.888).  The
near-parallel appearance is real on the log-odds scale --- POS multiplies
the odds of a correct gloss by a near-constant $\approx$1.8 --- and the
narrowing point gap is that constant advantage compressing near the
ceiling (see text).}
\label{fig:datasize}
\end{figure}

The interaction is substantial (Table~\ref{tab:datasize},
Figure~\ref{fig:datasize}): the gain peaks at $+11.6$ points with a quarter
of the data and shrinks to $+4.5$ with all of it --- a narrowing of $2.0$
points per doubling of training data (seed-level regression, interaction
$t=-5.0$; analysis released with our code).  At the smallest budget the
gain is somewhat smaller ($+9.5$ at 6 minutes than $+11.6$ at 12),
suggesting a floor below which the glossing model has too little data to
exploit the tier fully.  One qualification is important, because the two
curves in Figure~\ref{fig:datasize} nevertheless run roughly parallel to
the eye --- and on the \emph{log-odds} scale they are parallel: there the
interaction vanishes ($t=0.5$), and POS multiplies the odds of a correct
gloss by a factor that stays within 1.6--2.0 (mean $\approx$1.8) at every
budget.  The two readings are complementary, not contradictory.  POS
confers a roughly constant \emph{multiplicative} advantage; near the
accuracy ceiling that constant multiplier compresses into a few percentage
points, while in the low-data regime it unfolds into many.  What an
annotator experiences, however, is percentage points --- glosses to fix per
hundred --- so the practical return on the POS tier is largest exactly
where documentation projects begin.  The same fact can also be read
scale-free, as \emph{data
equivalence}: matching the POS-informed model trained on 78 utterances takes
the ablated model $\sim$221 utterances ($\times2.8$); at 155, $\sim$386
($\times2.5$); and the POS-informed model trained on \emph{half} the corpus
is not matched by the ablated model even with the \emph{full} corpus.  In
this range a POS tier more than halves the glossed data needed for a given
accuracy.  (These curves use gold POS and thus speak to the
assisted-annotation scenario; a pipeline analogue would retrain the tagger at
each size --- future work.)

\begin{figure*}[t]
\centering
\begin{minipage}[t]{0.475\textwidth}
\centering
\includegraphics[width=\linewidth]{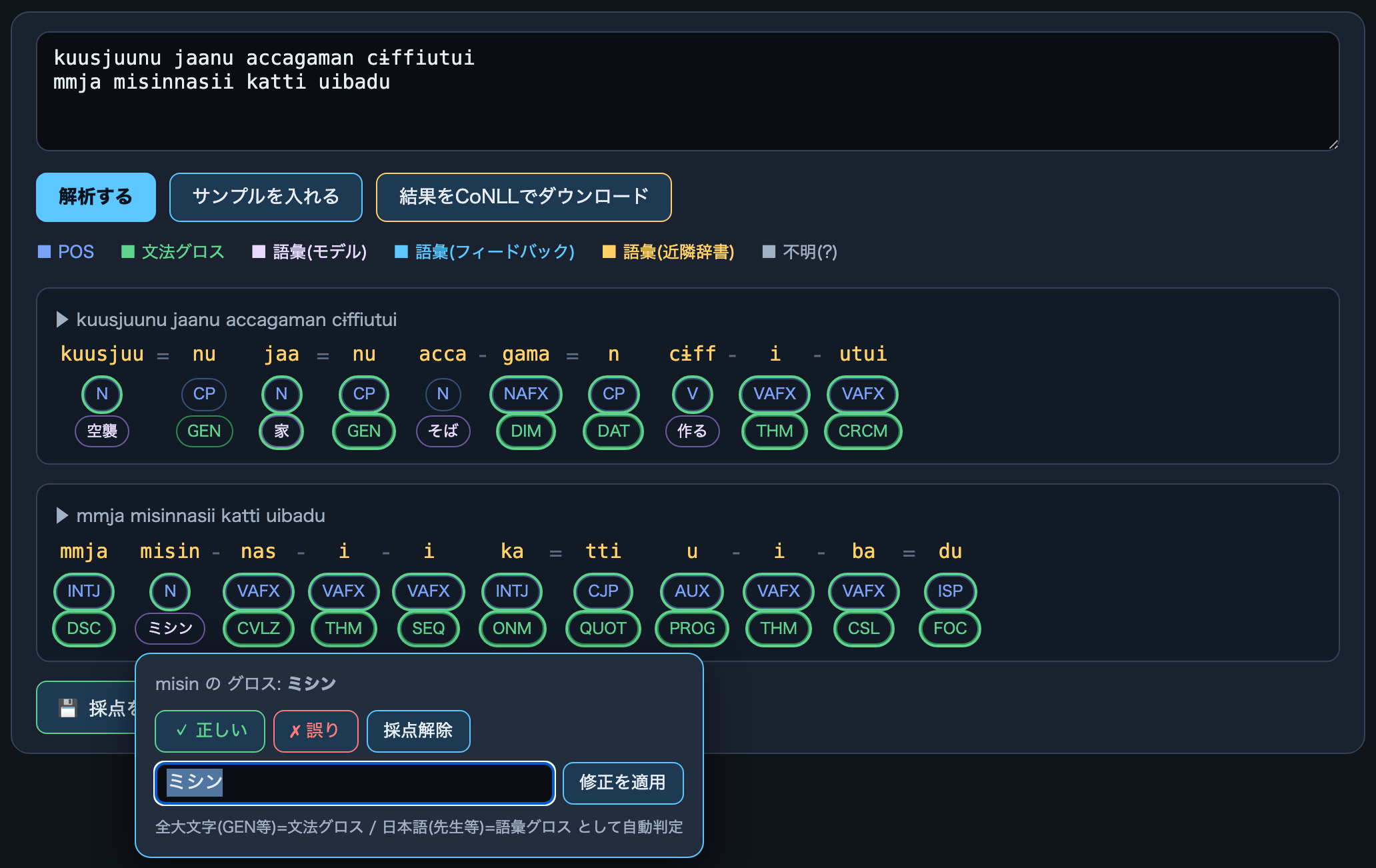}\\[2pt]
{\footnotesize (a) Interactive annotation view}
\end{minipage}\hfill
\begin{minipage}[t]{0.44\textwidth}
\centering
\includegraphics[width=\linewidth]{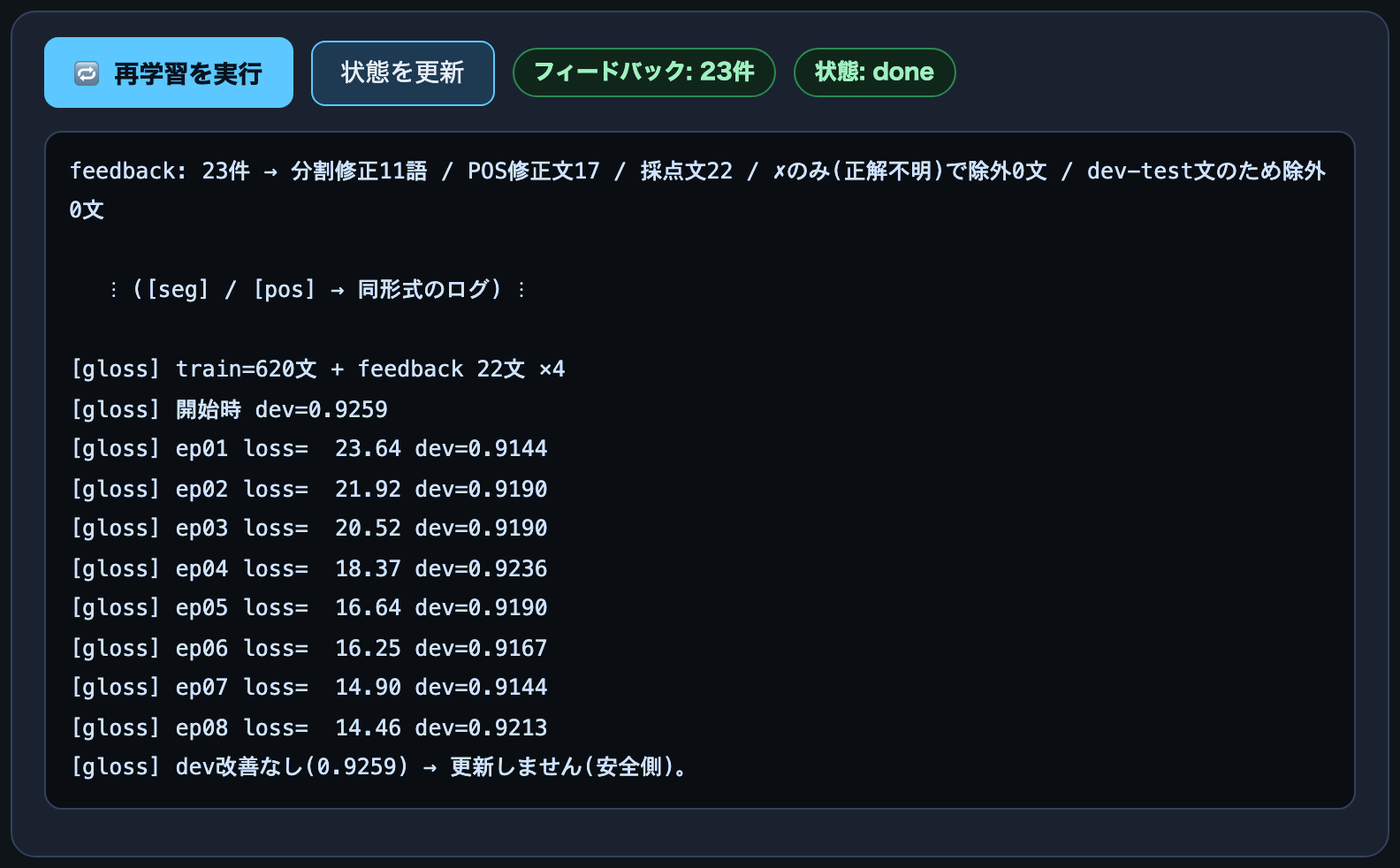}\\[2pt]
{\footnotesize (b) A fine-tuning run: per-epoch log with the dev gate}
\end{minipage}
\caption{Our Stage-2 annotation tool (browser interface; the
annotator-facing language is Japanese).  (a)~Free-text input,
interlinearized by the pipeline; every tier is confirmed or corrected in
place (green rings = confirmed tokens; the pop-up panel offers
confirm/reject/replace, and redrawing a word's morpheme boundaries
re-predicts its POS and glosses).  (b)~The log of an actual fine-tuning run
over 23 accumulated corrections (gloss section shown).  The header reports
the leak guard --- sentences coinciding with the held-out sets are excluded
(0 here) --- feedback is oversampled fourfold, and a model replaces the
deployed one only if its dev score improves; in this run the gate kept all
three incumbents.  Lexical-gloss corrections bypass retraining via a user
lexicon.}
\label{fig:tool}
\end{figure*}

\section{Implications for Annotation Practice}
\label{sec:implications}

Our results support a concrete recommendation for endangered-language
documentation: \emph{include a POS tier --- annotate quadrilinearly}.  The
argument has three layers, in decreasing order of immediacy.

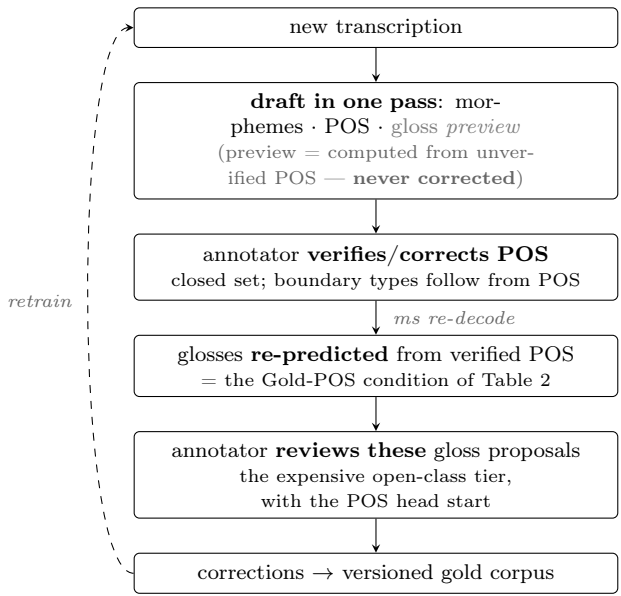
\begin{figure}[t]
\centering
\begin{tikzpicture}[>=stealth, font=\footnotesize, node distance=4.5mm,
  box/.style={draw, rounded corners=2pt, inner sep=4pt, align=center,
              text width=0.72\columnwidth},
  lab/.style={font=\scriptsize\itshape, color=black!60, align=left}]
  \node[box] (in) {new transcription};
  \node[box, below=of in] (draft) {\textbf{draft in one pass}: morphemes
    $\cdot$ POS $\cdot$ {\color{black!50}gloss \emph{preview}}\\
    {\scriptsize\color{black!60}(preview = computed from unverified POS ---
    \textbf{never corrected})}};
  \node[box, below=of draft] (pos) {annotator \textbf{verifies/corrects POS}\\
    {\scriptsize closed set; boundary types follow from POS}};
  \node[box, below=of pos] (re) {glosses \textbf{re-predicted} from verified POS\\
    {\scriptsize = the Gold-POS condition of Table~\ref{tab:main}}};
  \node[box, below=of re] (rev) {annotator \textbf{reviews these} gloss proposals\\
    {\scriptsize the expensive open-class tier, with the POS head start}};
  \node[box, below=of rev] (gold) {corrections $\rightarrow$ versioned gold corpus};
  \draw[->] (in) -- (draft);
  \draw[->] (draft) -- (pos);
  \draw[->] (pos) -- node[lab, right=1mm] {ms re-decode} (re);
  \draw[->] (re) -- (rev);
  \draw[->] (rev) -- (gold);
  \draw[->, dashed] (gold.west) .. controls +(-8mm,0) and +(-8mm,0) .. (in.west)
    node[lab, pos=0.5, left=1mm] {retrain};
\end{tikzpicture}
\caption{The assisted workflow, step by step.  The pipeline drafts all
tiers at once, but the draft's gloss line is a preview computed from
unverified POS and is never corrected directly.  The annotator verifies
POS; the glosses are re-predicted from the verified tags; and gloss review
happens only on these re-predictions --- the Gold-POS condition of
Table~\ref{tab:main}, with roughly 40\% fewer gloss errors to fix than
without a POS tier (10.7\%$\rightarrow$6.3\% at full data,
27.7\%$\rightarrow$16.1\% at 12 minutes).  Corrections flow into the
versioned gold corpus and, periodically, into retraining (dashed).}
\label{fig:assisted}
\end{figure}

\begin{enumerate}
\item \textbf{Computer-assisted annotation (immediate).}  In an interactive
  workflow where the annotator confirms or corrects POS before the model
  proposes glosses, the model effectively receives gold POS and the full
  $+4.35$-point gain applies.  Figure~\ref{fig:assisted} makes the order of
  operations explicit, because it is easy to misread: the pipeline drafts
  every tier at once, but the gloss line of that first draft is a
  \emph{preview}, computed from unverified POS --- it is never the object of
  correction.  The annotator verifies the POS line (boundaries follow
  automatically); the glosses are then \emph{re-predicted} from the verified
  tags --- a millisecond re-decode in our tool --- and only these proposals,
  now operating under the Gold-POS condition of Table~\ref{tab:main}, go to
  gloss review.  POS is a small closed set (32 tags) and far
  cheaper to annotate than glosses; a cheap tier accelerates the expensive
  one.  Note the cost asymmetry at the moment of annotation: for an
  annotator who has just segmented a morpheme and chosen its gloss, its
  part of speech is typically already decided in the analysis --- recording
  it is nearly free.  Conventional practice leaves this tier unwritten;
  our results show that the return on writing it down is large.  The data-size analysis (\S\ref{sec:datasize}) sharpens this: the gain
  is largest precisely when the corpus is small --- \emph{at the start of a
  documentation project} --- where a POS tier more than halves the glossed
  data needed.  The POS tier also subsumes another piece of IGT for free:
  the affix/clitic boundary typing (\emph{-} vs.\ \emph{=}) is recoverable
  from gold POS at 97.6\% (\S\ref{sec:pipeline}), so annotating POS
  effectively annotates boundary types as well.
\item \textbf{Fully automatic glossing (as annotation accumulates).}  The
  break-even analysis shows the pipeline gain is a function of tagger
  accuracy, which grows with annotated data.  Quadrilinear annotation is thus
  an investment: once the tagger clears roughly 90\%, POS begins to pay off
  automatically, approaching the $+4.35$-point ceiling.
\item \textbf{Model-side remedies (future work).}  The poisoning effect
  (Table~\ref{tab:positions}) suggests training with predicted or noised POS
  (scheduled sampling / POS dropout), joint POS--gloss modeling, or soft
  (probabilistic) POS inputs --- each attacks the $-2.32$-point term of
  Eq.~\eqref{eq:budget} rather than the tagger itself.
\end{enumerate}

We emphasize what our data do \emph{not} show: with the current corpus size,
quadrilinear annotation does not yet improve fully automatic glossing
significantly ($+0.40\pm0.77$).  The recommendation rests on the assisted
workflow, the measured ceiling, and the quantified path between them.

\section{Limitations}
\label{sec:limitations}
Our conclusions are bounded in five ways.  First, everything rests on a
single language and a single corpus --- roughly one hour of discourse, with a
test set of 405 grammatical tokens; the numbers should be read as one
carefully measured data point, not as universal constants.  Second, the
learning curves that ground the richness-beats-quantity argument use gold POS
at test time, so they speak directly to the computer-assisted scenario and
only indirectly to the fully automatic pipeline, whose tagger would itself
improve as data grow.  Third, we compare annotation designs under one
deliberately simple, fixed model family; we have not benchmarked multilingual
pretrained encoders or LLM-based glossing under the same annotation budget,
so our claims rank annotation strategies, not model families.  Fourth,
lexical glossing --- the open-class translation tier --- is treated only
peripherally, by lexicon copy with reranking.  Finally, the two-stage sizing
proposal (\S\ref{sec:conclusion}) extrapolates from one language's learning
curves, and its Stage-2 savings are a projection: the correction \emph{rate}
is measured, but the wall-clock review time in a live project is not yet.

\section{Conclusion: A Two-Stage Design for Discourse Documentation}
\label{sec:conclusion}

Discourse-collection projects for endangered languages routinely face a
sizing decision --- how much discourse to record, transcribe, and annotate ---
and, in our experience, that decision is usually made by convention (``one
hour,'' ``two hours'') rather than by evidence about what a given amount of
annotation buys.  Our budget-centered results turn this into an empirical
question, and they suggest a concrete answer for projects comparable to ours.

\paragraph{The proposal.}
Structure the project in \emph{two stages}:
\begin{enumerate}
\item \textbf{Stage 1 --- small, rich, careful.}  Collect roughly 30 minutes
  of discourse and invest the annotation effort there: full interlinearization
  in \emph{quadrilinear} format (text + POS + gloss + translation), verified
  with speakers.  Our results quantify what this buys: at a $\sim$23-minute
  training budget the POS-informed glossing model already reaches 0.89
  (assisted), and with our full $\sim$47-minute corpus the pipeline reaches
  segmentation span-F1 0.91, POS accuracy 0.88, and grammatical-gloss
  accuracy 0.93 when a human supplies POS (0.89--0.90 fully automatically).
\item \textbf{Stage 2 --- collect more, correct instead of create.}  For all
  subsequent recordings, let the Stage-1 pipeline draft the
  interlinearization, and shift human effort from \emph{creating} annotations
  to \emph{reviewing} them, in an interactive workflow where each correction
  feeds back into the models and the speaker-facing time is spent on
  verification, not mechanics.  We have built a working implementation of this
  workflow (Figure~\ref{fig:tool}): a browser-based tool in which the pipeline annotates arbitrary
  input, every tier (including morpheme boundaries) can be confirmed or
  corrected in place, lexical-gloss corrections take effect immediately
  through a user lexicon, and accumulated corrections periodically fine-tune
  the three models (oversampled relative to the base corpus).  Two design
  points matter for scientific hygiene: corrections whose sentences coincide
  with the held-out sets are automatically excluded from fine-tuning, so
  evaluation stays uncontaminated; and the deployed models are versioned
  separately from the frozen ones behind this paper's numbers.  The notorious
  60:1 transcription bottleneck becomes, for the annotation tiers, a review
  task --- and the review load is measurable: run end-to-end on our held-out
  test set, the pipeline delivers 63.6\% of word tokens fully correct across
  all tiers (187/294; 24 of the 77 utterances need no edit at all), so the
  reviewer edits roughly one word in three instead of authoring four tiers
  for every word.  Each reviewed hour, moreover, is not just annotated
  discourse but \emph{new training data}: folded back into the gold corpus
  as a versioned revision (with the hygiene above applied), it moves the
  models along the learning curves of \S\ref{sec:datasize} --- and once the
  tagger clears the break-even of \S\ref{sec:breakeven}, the currently
  latent POS gain compounds the improvement.  Review effort per hour of
  discourse thus falls as the archive grows, though not to zero: the curves
  flatten, and reviewed output must be audited before promotion to gold,
  lest model errors be ratified into the record.  How the correction rate
  converts into wall-clock savings is the Stage-2 measurement we call for
  below.
\end{enumerate}

\paragraph{Why 30 rich minutes rather than an hour of plain ones.}
The richness--quantity interaction (\S\ref{sec:datasize}) directly informs
how a fixed annotation budget should be split: a POS-informed model trained
on \emph{half} our corpus ($\sim$23 minutes) already matches an ablated model
trained on \emph{all} of it ($\sim$47 minutes; 0.890 vs.\ 0.888) --- rich
half-hour-scale annotation buys what plain hour-scale annotation buys.  Since the POS tier adds only a small
fraction of the per-utterance cost --- a closed inventory
of 32 tags, versus
open-ended glossing and translation --- richly annotating less discourse is
the better investment than thinly annotating more.  In the low-data regime
where every documentation project begins, annotation \emph{richness beats
quantity}.

\paragraph{A concrete horizon: legacy discourse archives.}
The scale of the opportunity in Japan alone is illustrated by the Agency for
Cultural Affairs' emergency dialect survey (1977--1985), which recorded
roughly ten hours of discourse at each of 228 sites nationwide.  Four
decades later, making those recordings research-ready --- digitization,
alignment, native checking --- is still ongoing, sustained by collaborative
projects that recruit volunteer researchers and distribute the manual labor
\citep{hidaka2026}.  Legacy archives of this kind are precisely where the
two-stage design points: for any variety in the archive that a researcher
commands, a Stage-1 investment of some thirty richly annotated minutes would
let a pipeline draft the remaining hours for \emph{review} rather than
creation.  The current curation efforts target transcription and phrase-level
translation rather than IGT; our proposal complements them by showing how
such archives could be lifted to the IGT standard at realistic cost --- and
the two efforts could share infrastructure, since curated, aligned
transcripts are exactly the input our pipeline expects.

\paragraph{Outlook.}
The specific figure of 30 minutes is derived from one language and one
corpus; the two-stage logic, however, only requires that the learning curves
retain their shape --- steep early gains, richness amplifying them --- which we
expect to be robust for agglutinative languages with comparable morphological
transparency.  Replicating the budget analysis across languages, closing the
gap between oracle and pipeline POS (via better taggers or POS-robust
training), and measuring the actual review-time savings in Stage~2 are the
natural next steps.

\bibliography{references}

\clearpage
\appendix

\section{POS Tag Inventory}
\label{app:tagset}

Table~\ref{tab:tagset} lists the complete POS tag inventory with corpus
frequencies.  Tag names follow the annotation scheme of the Irabu reference
grammar \citep{shimoji2017}.  The inventory has a long tail: five tags occur
exactly once in the corpus, and \textsc{onm} occurs only outside the
training split --- the source of the unseen-label errors visible in
Table~\ref{tab:perlabel}.

\begin{table}[H]
\centering
\footnotesize
\setlength{\tabcolsep}{3.5pt}
\begin{tabular}{lrr@{\quad}lrr}
\toprule
Tag & Train & Total & Tag & Train & Total \\
\midrule
\textsc{vafx} & 1058 & 1328 & \textsc{fn}   & 45 & 58 \\
\textsc{n}    &  699 &  862 & \textsc{pn}   & 43 & 55 \\
\textsc{v}    &  609 &  763 & \textsc{xafx} & 34 & 43 \\
\textsc{cp}   &  540 &  667 & \textsc{adn}  & 30 & 36 \\
\textsc{isp}  &  344 &  460 & \textsc{num}  & 28 & 34 \\
\textsc{intj} &  329 &  404 & \textsc{unk}  & 19 & 25 \\
\textsc{prn}  &  195 &  249 & \textsc{aafx} & 19 & 22 \\
\textsc{sfp}  &  193 &  236 & \textsc{clf}  & 16 & 20 \\
\textsc{aux}  &  158 &  212 & \textsc{mp}   & 10 & 13 \\
\textsc{jpn}  &  148 &  199 & \textsc{rdb}  &  3 &  7 \\
\textsc{cjp}  &  158 &  197 & \textsc{onm}  &  0 &  2 \\
\textsc{lp}   &  112 &  143 & \textsc{dat}  &  1 &  1 \\
\textsc{nafx} &  101 &  140 & \textsc{dem}  &  1 &  1 \\
\textsc{conj} &   67 &   92 & \textsc{pst}  &  1 &  1 \\
\textsc{adj}  &   60 &   74 & \textsc{comj} &  1 &  1 \\
\textsc{adv}  &   52 &   66 & \textsc{top}  &  1 &  1 \\
\bottomrule
\end{tabular}
\caption{POS tag inventory with token frequencies (training split /
whole corpus; 5{,}075 and 6{,}412 tokens respectively).}
\label{tab:tagset}
\end{table}

\section{Per-Label Performance}
\label{app:perlabel}

Table~\ref{tab:perlabel} breaks the POS tagger's test performance down by
tag, under exactly the condition behind the headline number of
\S\ref{sec:pipeline} (sentence-by-sentence decoding; overall accuracy
$583/662=0.881$).  Table~\ref{tab:glosslabel} does the same for grammatical
glossing, comparing the No-POS and Gold-POS conditions of
\S\ref{sec:results} label by label (5-seed means).  The gains concentrate
where \S\ref{sec:results} locates them --- case marking
(\gascore{nom} $+14.1$, \gascore{dat} $+16.4$, \gascore{acc} $+10.6$) and
the focus/topic clitics (\gascore{foc} $+3.6$, \gascore{top} $+3.3$) ---
while already-unambiguous labels (\gascore{pst}, \gascore{quot},
\gascore{prog}) sit at ceiling in both conditions.

\begin{table}[H]
\centering
\footnotesize
\setlength{\tabcolsep}{4pt}
\begin{tabular}{lrccc}
\toprule
Tag & Support & Recall & Precision & F1 \\
\midrule
\textsc{vafx} & 136 & 0.993 & 0.938 & 0.964 \\
\textsc{n}    &  85 & 0.812 & 0.802 & 0.807 \\
\textsc{v}    &  78 & 0.910 & 0.845 & 0.877 \\
\textsc{cp}   &  70 & 0.929 & 0.915 & 0.922 \\
\textsc{isp}  &  55 & 0.945 & 0.981 & 0.963 \\
\textsc{intj} &  32 & 0.812 & 0.929 & 0.867 \\
\textsc{jpn}  &  31 & 0.871 & 0.794 & 0.831 \\
\textsc{aux}  &  25 & 0.800 & 0.833 & 0.816 \\
\textsc{nafx} &  23 & 0.826 & 0.950 & 0.884 \\
\textsc{prn}  &  21 & 0.952 & 0.952 & 0.952 \\
\textsc{cjp}  &  19 & 0.947 & 0.900 & 0.923 \\
\textsc{sfp}  &  16 & 0.750 & 0.750 & 0.750 \\
\textsc{conj} &  15 & 0.933 & 1.000 & 0.966 \\
\textsc{lp}   &  12 & 0.833 & 0.714 & 0.769 \\
\textsc{adj}  &   7 & 0.286 & 1.000 & 0.444 \\
\textsc{fn}   &   6 & 0.667 & 0.571 & 0.615 \\
\textsc{pn}   &   6 & 0.500 & 1.000 & 0.667 \\
\textsc{adv}  &   6 & 0.667 & 0.667 & 0.667 \\
\textsc{xafx} &   4 & 1.000 & 0.667 & 0.800 \\
\textsc{unk}  &   3 & 0.667 & 0.667 & 0.667 \\
\textsc{mp}   &   3 & 0.667 & 1.000 & 0.800 \\
\textsc{onm}  &   2 & 0.000 & 0.000 & 0.000 \\
\textsc{rdb}  &   2 & 1.000 & 1.000 & 1.000 \\
\textsc{num}  &   2 & 0.500 & 1.000 & 0.667 \\
\textsc{adn}  &   1 & 0.000 & 0.000 & 0.000 \\
\textsc{aafx} &   1 & 0.000 & 0.000 & 0.000 \\
\textsc{clf}  &   1 & 1.000 & 1.000 & 1.000 \\
\bottomrule
\end{tabular}
\caption{POS tagger, per-tag test results (662 tokens, overall accuracy
0.881).  \textsc{onm} does not occur in the training data.}
\label{tab:perlabel}
\end{table}

\begin{table}[H]
\centering
\footnotesize
\setlength{\tabcolsep}{3pt}
\begin{tabular}{lrccr}
\toprule
Gloss & Supp. & No POS & Gold POS & $\Delta$ (pts) \\
\midrule
\gascore{thm}  & 45 & 0.996 & 1.000 & $+0.4$ \\
\gascore{seq}  & 29 & 0.931 & 0.924 & $-0.7$ \\
\gascore{foc}  & 28 & 0.964 & 1.000 & $+3.6$ \\
\gascore{top}  & 24 & 0.967 & 1.000 & $+3.3$ \\
\gascore{pst}  & 20 & 1.000 & 1.000 & $+0.0$ \\
\gascore{acc}  & 17 & 0.894 & 1.000 & $+10.6$ \\
\gascore{nom}  & 17 & 0.624 & 0.765 & $+14.1$ \\
\gascore{gen}  & 15 & 0.960 & 0.973 & $+1.3$ \\
\gascore{quot} & 15 & 1.000 & 1.000 & $+0.0$ \\
\gascore{dsc}  & 14 & 0.986 & 1.000 & $+1.4$ \\
\gascore{csl}  & 14 & 0.971 & 1.000 & $+2.9$ \\
\gascore{prog} & 12 & 1.000 & 1.000 & $+0.0$ \\
\gascore{dat}  & 11 & 0.836 & 1.000 & $+16.4$ \\
\gascore{pl}   &  9 & 0.956 & 1.000 & $+4.4$ \\
\gascore{dim}  &  9 & 1.000 & 1.000 & $+0.0$ \\
\gascore{cnf}  &  8 & 0.850 & 0.900 & $+5.0$ \\
\gascore{3sg}  &  8 & 0.850 & 1.000 & $+15.0$ \\
\gascore{add}  &  8 & 1.000 & 1.000 & $+0.0$ \\
\gascore{cop}  &  6 & 0.967 & 0.967 & $+0.0$ \\
\gascore{onm}  &  5 & 0.040 & 0.320 & $+28.0$ \\
\gascore{rls}  &  5 & 1.000 & 1.000 & $+0.0$ \\
\midrule
(others, pooled) & 86 & 0.765 & 0.830 & $+6.5$ \\
\bottomrule
\end{tabular}
\caption{Grammatical glossing, per-label test accuracy under the No-POS and
Gold-POS conditions (5-seed means; labels with at least 5 test tokens,
remainder pooled as ``others'').  The POS gain concentrates on case marking and
focus/topic clitics.}
\label{tab:glosslabel}
\end{table}

\end{document}